\documentclass[twoside,11pt]{article}

\usepackage{jmlr2e}

\usepackage[normalem]{ulem}

\usepackage{tikz}
\usetikzlibrary{tikzmark,calc,fit}
\usepackage{pythonhighlight}
\usepackage{ifthen}
\newboolean{draft}
\setboolean{draft}{true}
\ifthenelse{\boolean{draft}}{%
  \newcounter{comments}
  \newcommand{\jamie}[1]{\addtocounter{comments}{1}{\color{Tomato}[JB \thecomments: #1]}}
  \newcommand{\richard}[1]{\addtocounter{comments}{1}{\color{DarkGreen}[YCL \thecomments: #1]}}
  \newcommand{\gaurav}[1]{\addtocounter{comments}{1}{\color{DarkRed}[GP \thecomments: #1]}}
  
}{
\newcommand{\jamie}[1]{}
\newcommand{\richard}[1]{}
\newcommand{\gaurav}[1]{}

}

\ShortHeadings{}{}
\firstpageno{1}

\begin{document}

\title{eipy: An Open-Source Python Package for Multi-modal Data Integration using Heterogeneous Ensembles}

\author{\name Jamie J. R. Bennett \email jamie.bennett@mssm.edu 
       \AND
       \name Aviad Susman \email aviad.susman@mssm.edu 
       \AND
       \name Yan Chak Li \email yan-chak.li@mssm.edu 
       \AND
       \name Gaurav Pandey \email gaurav.pandey@mssm.edu \\
       \addr Department of Genetics and Genomic Sciences,\\
       Icahn School of Medicine at Mount Sinai, \\
       New York, NY 10029, USA
}

\editor{}

\maketitle

\begin{abstract}%   <- trailing '%' for backward compatibility of .sty file
In this paper, we introduce eipy--an open-source Python package for developing effective, multi-modal heterogeneous ensembles for classification. eipy simultaneously provides both a rigorous, and user-friendly framework for comparing and selecting the best-performing multi-modal data integration and predictive modeling methods by systematically evaluating their performance using nested cross-validation. The package is designed to leverage scikit-learn-like estimators as components to build multi-modal predictive models. An up-to-date user guide, including API reference and tutorials, for eipy is maintained at https://eipy.readthedocs.io. The main repository for this project can be found on GitHub at https://github.com/GauravPandeyLab/eipy.  
\end{abstract}


\section{Introduction}
Multi-modal data combine diverse information from multiple sources or modalities, and offer the ability to make novel findings and predictions about the entities described \citep{boehm2022harnessing, krassowski2020state}. The rapid increase in the availability of these data has led to a surge in multi-modal machine learning methods and applications \citep{yoon_jpm_2023, bajpai_ieee_2023, xing_renewableenergy_2023}. Effective utilization of the complementary information across the constituent data modalities to maximize the predictive power of methods is therefore an important goal for machine learning research.

Recent software packages have been proposed as an attempt to automate the building of machine learning models from multi-modal data. \verb|MULTIZOO| \citep{liang_2023_multizoo} focuses on automating the building of multi-modal deep learning models. However, it has been noted that deep learning models are particularly difficult to train in the multi-modal setting \citep{wang_2020}, and their uni-modal counterpart models can often outperform them, impacting its usability for non-specialists. \verb|scikit-multimodallearn| \citep{benielli_2022_toolbox} provides a scikit-learn style API for building multi-modal prediction models, which can be incorporated in the \verb|scikit-learn| pipeline. A selection of boosting and kernel based algorithms are provided, however, data inputs must be structured and there is no scope for the inclusion of unstructured data modalities in future releases. Furthermore, these software packages are not focused on comparing and evaluating different methods of multi-modal data integration.

The \verb|eipy| software presented in this paper implements Ensemble Integration (EI), a framework for multi-modal predictive modeling \citep{li_bioinformatics_2022}. Specifically, EI builds on the theoretical foundations of stacked generalization \citep{wolpert_neuralnetworks_1992} by utilizing heterogeneous ensembles as a data integration method. On top of this, EI automates a nested-cross validation procedure to ensure rigorous and fair training and evaluation of user-defined base/ensemble prediction methods. The original motivation for developing such an algorithm was to build towards precision medicine \citep{hodson_nature_2016} by utilizing the abundance of multi-modal bio-medical and healthcare data \citep{topol_science_2023, kline_npj_multimodal}. We demonstrated the benefits of EI in several biomedical applications, including the prediction of mortality associated with COVID-19 \citep{li_bioinformatics_2022}, protein function prediction \citep{li_bioinformatics_2022} and the detection of diabetes among youth \citep{mcdonough_2023}. However, the general-purpose design of the EI framework can enable its use in a wide range of multi-modal applications in other areas as well. For example, sentiment analysis can combine various modalities, including words, visual expression and acoustics \citep{bagher-2018-multimodal}. Once features are extracted from each of these modalities, EI could then be used to fuse the data and produce predictive models. Another potential application could be classification of remote sensing data \citep{wu_2022_remotesensing} that includes multiple types of imaging modalities relating to the same location.

\begin{figure}[t!]
\centering
\makebox[0pt][c]{\includegraphics[scale=1]{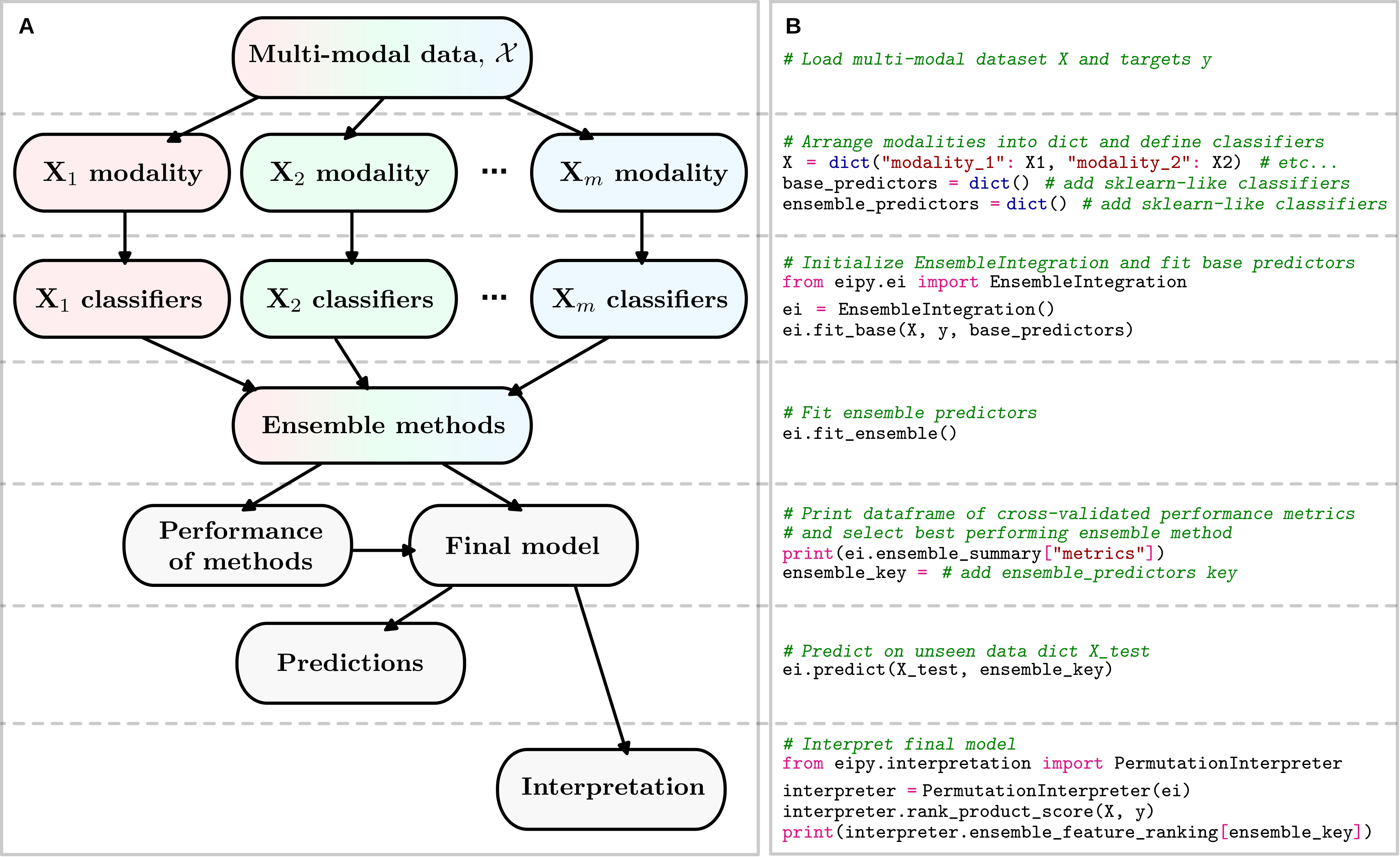}}
\caption{Overview of Ensemble Integration (EI) and the eipy API. A. Schematic illustration of the EI workflow. B. The corresponding eipy commands to implement each step of EI. } \label{fig:eipy}
\end{figure}

\section{Ensemble Integration}
Consider multi-modal datasets of the form $\mathcal{X} = \left\{ \mathbf{X}_i \in \mathbb{R}^{n \times f_i } \, | \, i = 1, 2, \ldots, m \right\}$, where $m$ is the number of modalities, $n$ is the number of samples, and $f_i$ is the number of features in modality $i$. The general setting of heterogeneous ensembles (see for example \cite{reid2007review}) can be employed for multi-modal datasets relatively easily (Fig. \ref{fig:eipy}A). We can train a unique set of base predictors on each modality $X_i$, and base prediction vectors can be combined across all modalities to train the ensembles via nested CV \citep{li_bioinformatics_2022, whalen_2016}. Heterogeneity in the ensembles now not only comes from differences in base predictor algorithms, but also via the modalities they have been trained on.

In addition to their prediction capabilities, a key feature of EI models is that they can be interpreted. The original EI framework \citep{li_bioinformatics_2022} includes a permutation-based approach to interpret the final models, and highlight features across the input modalities that contribute the most to the models’ predictions. This interpreter combines performance importance scores \citep{breiman_2001_random} at base- and ensemble- predictor stage to obtain a final aggregated feature contribution score and rank for each feature. 

A more in depth description of the EI framework can be found in the original methods paper \citep{li_bioinformatics_2022}.

\section{Software overview}
Implementing the EI process for each use case is complicated, especially for non-specialists. eipy provides an easy-to-use framework to build interpretable, multi-modal ensembles by automating the nested cross-validation procedure. eipy’s API (Fig. 1B) is designed to be similar to that of scikit-learn \citep{scikit-learn}. \verb|EnsembleIntegration| is the main class that handles all performance analysis of ensemble methods, as well as fitting the final model. 

\texttt{EnsembleIntegration} implements three main methods; firstly, \verb|fit_base| trains the base predictors on each modality $X_i$, and generates ensemble training data for the heterogeneous ensembles. It can be called on a dictionary containing all $X_i$, as shown in Fig. 1B, or it can be called iteratively on each $X_i$ individually. The latter method is more flexible, allowing one to fit a specific set of base predictors to each modality. Once base predictors have been trained, and ensemble training data generated, \verb|fit_ensemble| can be called to train the ensemble predictors. Performance of both base- and ensemble-predictors can be evaluated using the \verb|base_summary| and \verb|ensemble_summary| attributes. The best performing ensemble can then be selected, and predictions with this model can be made with \verb|predict|. In addition, we provide the \verb|PermutationInterpreter| class for the interpretation of the multi-modal heterogeneous ensembles built using \verb|eipy|. The class acts as a wrapper for the (fitted) \verb|EnsembleIntegration| object. 

Finally, we have provided a pre-processed multi-modal youth diabetes dataset \citep{mcdonough_2023}. This epidemiological dataset was extracted from the  U.S. National Health and Nutrition Examination Survey (NHANES). It covered more than 15000 youth US participants with 95 potentially diabetes-related variables grouped by 4 modalities: sociodemographics, health status, diet and other lifestyle behavior. This inclusion provides users an example of a real world, multi-modal biomedical dataset, which can be loaded with the \verb|load_diabetes| function.

\section{Development}
To optimize project development and long-term maintenance, we employed several software development best practices. We used Poetry \citep{poetry} to manage package dependencies more easily. The codebase adheres to high-quality standards through continuous linting using \verb|flake8| \citep{flake8} and consistent formatting with \verb|black| \citep{black}. To facilitate easy installation, we have published the package to PyPI, so that one can install with pip. 

We maintain version control using GitHub, enabling collaborative development and transparent tracking of changes. We employed continuous integration workflows via GitHub Actions to automate high-coverage testing with \verb|pytest| \citep{pytest}. In addition, compatibility testing was conducted across multiple operating systems and Python versions with \verb|tox| \citep{tox}, to ensure broad accessibility. 

We have made the project open-source and accessible to the community. To this end, we prioritized documentation, utilizing \verb|sphinx| \citep{sphinx} and Read the Docs to provide comprehensive auto-generated online documentation. Additionally, we offer clear development instructions, allowing contributors to participate actively in the project’s growth and development.

\section{Future work}
Advancement of \verb|eipy| will focus on a few areas. 
\begin{itemize}
    \item We will extend the codebase to include functionalities to conduct multi-class and multi-label classification, as well as regression-type tasks in future releases. 
    \item Multi-modal data can have a mix of structured and/or unstructured modalities. Currently, for unstructured data, feature extraction must be performed prior to input into \verb|eipy|. We aim to include deep learning methods as base predictors in \verb|eipy|, so that unstructured data modalities can be better utilized in their native form. 
    \item We will enable model interpretation via Shapley values \citep{lundberg_2017, chen_ncomms_2022}, which have shown promise in numerous studies \citep{arrieta_2020, zoabi_2021, lundberg_2020}. 
    \item We will include more pre-processed multi-modal datasets to enable more extensive testing and utilization of eipy, as well as other multi-modal methods.
\end{itemize}
With these advances, eipy will become applicable and useful to an even broader range of multi-modal datasets and prediction problems.

\section*{Acknowledgements}

We would like to thank Catherine McDonough, Nita Vangeepuram and Bian Liu for preparing and allowing us to publish the multi-modal diabetes dataset with eipy. The development and testing process of eipy was run in part by computational resources provided by Scientific Computing at the Icahn School of Medicine at Mount Sinai, Oracle Cloud credits and related resources provided by the Oracle for Research program.
This work was funded by NIH grant R01HG011407.

\vskip 0.2in
\bibliography{main}

\begin{thebibliography}{31}
\providecommand{\natexlab}[1]{#1}
\providecommand{\url}[1]{\texttt{#1}}
\expandafter\ifx\csname urlstyle\endcsname\relax
  \providecommand{\doi}[1]{doi: #1}\else
  \providecommand{\doi}{doi: \begingroup \urlstyle{rm}\Url}\fi

\bibitem[Arrieta et~al.(2020)Arrieta, D{\'\i}az-Rodr{\'\i}guez, Del~Ser, Bennetot, Tabik, Barbado, Garc{\'\i}a, Gil-L{\'o}pez, Molina, Benjamins, et~al.]{arrieta_2020}
Alejandro~Barredo Arrieta, Natalia D{\'\i}az-Rodr{\'\i}guez, Javier Del~Ser, Adrien Bennetot, Siham Tabik, Alberto Barbado, Salvador Garc{\'\i}a, Sergio Gil-L{\'o}pez, Daniel Molina, Richard Benjamins, et~al.
\newblock Explainable artificial intelligence (xai): Concepts, taxonomies, opportunities and challenges toward responsible ai.
\newblock \emph{Information fusion}, 58:\penalty0 82--115, 2020.

\bibitem[Bagher~Zadeh et~al.(2018)Bagher~Zadeh, Liang, Poria, Cambria, and Morency]{bagher-2018-multimodal}
AmirAli Bagher~Zadeh, Paul~Pu Liang, Soujanya Poria, Erik Cambria, and Louis-Philippe Morency.
\newblock Multimodal language analysis in the wild: {CMU}-{MOSEI} dataset and interpretable dynamic fusion graph.
\newblock In \emph{Proceedings of the 56th Annual Meeting of the Association for Computational Linguistics (Volume 1: Long Papers)}, pages 2236--2246, Melbourne, Australia, July 2018. Association for Computational Linguistics.
\newblock \doi{10.18653/v1/P18-1208}.
\newblock URL \url{https://aclanthology.org/P18-1208}.

\bibitem[Bajpai et~al.(2023)Bajpai, Khare, and Joshi]{bajpai_ieee_2023}
Rishabh Bajpai, Suyash Khare, and Deepak Joshi.
\newblock A multi-modal model-fusion approach for improved prediction of freezing of gait in parkinson’s disease.
\newblock \emph{IEEE Sensors Journal}, 2023.

\bibitem[Benielli et~al.(2022)Benielli, Capponi, Ko{\c{c}}o, Kadri, Huusari, Bauvin, and Laviolette]{benielli_2022_toolbox}
Dominique Benielli, C{\'e}cile Capponi, Sokol Ko{\c{c}}o, Hachem Kadri, Riikka Huusari, Baptiste Bauvin, and Fran{\c{c}}ois Laviolette.
\newblock Toolbox for multimodal learn (scikit-multimodallearn).
\newblock \emph{The Journal of Machine Learning Research}, 23\penalty0 (1):\penalty0 2407--2413, 2022.

\bibitem[Boehm et~al.(2022)Boehm, Khosravi, Vanguri, Gao, and Shah]{boehm2022harnessing}
Kevin~M Boehm, Pegah Khosravi, Rami Vanguri, Jianjiong Gao, and Sohrab~P Shah.
\newblock Harnessing multimodal data integration to advance precision oncology.
\newblock \emph{Nature Reviews Cancer}, 22\penalty0 (2):\penalty0 114--126, 2022.

\bibitem[Breiman(2001)]{breiman_2001_random}
Leo Breiman.
\newblock Random forests.
\newblock \emph{Machine learning}, 45:\penalty0 5--32, 2001.

\bibitem[Chen et~al.(2022)Chen, Lundberg, and Lee]{chen_ncomms_2022}
Hugh Chen, Scott~M Lundberg, and Su-In Lee.
\newblock Explaining a series of models by propagating shapley values.
\newblock \emph{Nature communications}, 13\penalty0 (1):\penalty0 4512, 2022.

\bibitem[Eustace et~al.(2023)]{poetry}
Sébastien Eustace et~al.
\newblock Poetry: Python packaging and dependency management made easy, 2023.
\newblock URL \url{https://python-poetry.org/}.

\bibitem[Hodson(2016)]{hodson_nature_2016}
Richard Hodson.
\newblock Precision medicine.
\newblock \emph{Nature}, 537\penalty0 (7619):\penalty0 S49--S49, 2016.

\bibitem[Kline et~al.(2022)Kline, Wang, Li, Dennis, Hutch, Xu, Wang, Cheng, and Luo]{kline_npj_multimodal}
Adrienne Kline, Hanyin Wang, Yikuan Li, Saya Dennis, Meghan Hutch, Zhenxing Xu, Fei Wang, Feixiong Cheng, and Yuan Luo.
\newblock Multimodal machine learning in precision health: A scoping review.
\newblock \emph{npj Digital Medicine}, 5\penalty0 (1):\penalty0 171, 2022.

\bibitem[Krassowski et~al.(2020)Krassowski, Das, Sahu, and Misra]{krassowski2020state}
Michal Krassowski, Vivek Das, Sangram~K Sahu, and Biswapriya~B Misra.
\newblock State of the field in multi-omics research: from computational needs to data mining and sharing.
\newblock \emph{Frontiers in Genetics}, 11:\penalty0 610798, 2020.

\bibitem[Krekel et~al.(2023{\natexlab{a}})Krekel, Oliveira, Pfannschmidt, Bruynooghe, Laugher, and Bruhin]{pytest}
Holger Krekel, Bruno Oliveira, Ronny Pfannschmidt, Floris Bruynooghe, Brianna Laugher, and Florian Bruhin.
\newblock pytest: helps you write better programs, 2023{\natexlab{a}}.
\newblock URL \url{https://docs.pytest.org/}.

\bibitem[Krekel et~al.(2023{\natexlab{b}})]{tox}
Holger Krekel et~al.
\newblock tox - automation project, 2023{\natexlab{b}}.
\newblock URL \url{https://tox.wiki/}.

\bibitem[Langa et~al.(2023)]{black}
Łukasz Langa et~al.
\newblock Black: The uncompromising python code formatter, 2023.
\newblock URL \url{https://black.readthedocs.io/en/stable/}.

\bibitem[Li et~al.(2022)Li, Wang, Law, Murali, and Pandey]{li_bioinformatics_2022}
Yan~Chak Li, Linhua Wang, Jeffrey~N Law, T~M Murali, and Gaurav Pandey.
\newblock {Integrating multimodal data through interpretable heterogeneous ensembles}.
\newblock \emph{Bioinformatics Advances}, 2\penalty0 (1):\penalty0 vbac065, 09 2022.
\newblock ISSN 2635-0041.
\newblock \doi{10.1093/bioadv/vbac065}.
\newblock URL \url{https://doi.org/10.1093/bioadv/vbac065}.

\bibitem[Liang et~al.(2023)Liang, Lyu, Fan, Agarwal, Cheng, Morency, and Salakhutdinov]{liang_2023_multizoo}
Paul~Pu Liang, Yiwei Lyu, Xiang Fan, Arav Agarwal, Yun Cheng, Louis-Philippe Morency, and Ruslan Salakhutdinov.
\newblock Multizoo \& multibench: A standardized toolkit for multimodal deep learning.
\newblock \emph{Journal of Machine Learning Research}, 24:\penalty0 1--7, 2023.

\bibitem[Lundberg and Lee(2017)]{lundberg_2017}
Scott~M Lundberg and Su-In Lee.
\newblock A unified approach to interpreting model predictions.
\newblock In I.~Guyon, U.~Von Luxburg, S.~Bengio, H.~Wallach, R.~Fergus, S.~Vishwanathan, and R.~Garnett, editors, \emph{Advances in Neural Information Processing Systems}, volume~30. Curran Associates, Inc., 2017.
\newblock URL \url{https://proceedings.neurips.cc/paper_files/paper/2017/file/8a20a8621978632d76c43dfd28b67767-Paper.pdf}.

\bibitem[Lundberg et~al.(2020)Lundberg, Erion, Chen, DeGrave, Prutkin, Nair, Katz, Himmelfarb, Bansal, and Lee]{lundberg_2020}
Scott~M Lundberg, Gabriel Erion, Hugh Chen, Alex DeGrave, Jordan~M Prutkin, Bala Nair, Ronit Katz, Jonathan Himmelfarb, Nisha Bansal, and Su-In Lee.
\newblock From local explanations to global understanding with explainable ai for trees.
\newblock \emph{Nature machine intelligence}, 2\penalty0 (1):\penalty0 56--67, 2020.

\bibitem[McDonough et~al.(2023)McDonough, Li, Vangeepuram, Liu, and Pandey]{mcdonough_2023}
Catherine McDonough, Yan~Chak Li, Nita Vangeepuram, Bian Liu, and Gaurav Pandey.
\newblock Facilitating youth diabetes studies with the most comprehensive epidemiological dataset available through a public web portal.
\newblock \emph{medRxiv}, 2023.
\newblock \doi{10.1101/2023.08.02.23293517}.
\newblock URL \url{https://www.medrxiv.org/content/early/2023/08/04/2023.08.02.23293517}.

\bibitem[Pedregosa et~al.(2011)Pedregosa, Varoquaux, Gramfort, Michel, Thirion, Grisel, Blondel, Prettenhofer, Weiss, Dubourg, Vanderplas, Passos, Cournapeau, Brucher, Perrot, and Duchesnay]{scikit-learn}
F.~Pedregosa, G.~Varoquaux, A.~Gramfort, V.~Michel, B.~Thirion, O.~Grisel, M.~Blondel, P.~Prettenhofer, R.~Weiss, V.~Dubourg, J.~Vanderplas, A.~Passos, D.~Cournapeau, M.~Brucher, M.~Perrot, and E.~Duchesnay.
\newblock Scikit-learn: Machine learning in {P}ython.
\newblock \emph{Journal of Machine Learning Research}, 12:\penalty0 2825--2830, 2011.

\bibitem[Reid(2007)]{reid2007review}
Sam Reid.
\newblock A review of heterogeneous ensemble methods.
\newblock \emph{Department of Computer Science, University of Colorado at Boulder}, 2007.

\bibitem[Topol(2023)]{topol_science_2023}
Eric~J. Topol.
\newblock As artificial intelligence goes multimodal, medical applications multiply.
\newblock \emph{Science}, 381\penalty0 (6663):\penalty0 eadk6139, 2023.
\newblock \doi{10.1126/science.adk6139}.
\newblock URL \url{https://www.science.org/doi/abs/10.1126/science.adk6139}.

\bibitem[Turner et~al.(2023)Turner, Ronacher, and Neuhäuser]{sphinx}
Adam Turner, Armin Ronacher, and Daniel Neuhäuser.
\newblock Sphinx, 2023.
\newblock URL \url{https://www.sphinx-doc.org/}.

\bibitem[Wang et~al.(2020)Wang, Tran, and Feiszli]{wang_2020}
Weiyao Wang, Du~Tran, and Matt Feiszli.
\newblock What makes training multi-modal classification networks hard?
\newblock In \emph{Proceedings of the IEEE/CVF conference on computer vision and pattern recognition}, pages 12695--12705, 2020.

\bibitem[Whalen et~al.(2016)Whalen, Pandey, and Pandey]{whalen_2016}
Sean Whalen, Om~Prakash Pandey, and Gaurav Pandey.
\newblock Predicting protein function and other biomedical characteristics with heterogeneous ensembles.
\newblock \emph{Methods}, 93:\penalty0 92--102, 2016.
\newblock ISSN 1046-2023.
\newblock \doi{https://doi.org/10.1016/j.ymeth.2015.08.016}.
\newblock URL \url{https://www.sciencedirect.com/science/article/pii/S1046202315300566}.
\newblock Computational protein function predictions.

\bibitem[Wolpert(1992)]{wolpert_neuralnetworks_1992}
David~H. Wolpert.
\newblock Stacked generalization.
\newblock \emph{Neural Networks}, 5\penalty0 (2):\penalty0 241--259, 1992.
\newblock ISSN 0893-6080.
\newblock \doi{https://doi.org/10.1016/S0893-6080(05)80023-1}.
\newblock URL \url{https://www.sciencedirect.com/science/article/pii/S0893608005800231}.

\bibitem[Wu et~al.(2022)Wu, Hong, and Chanussot]{wu_2022_remotesensing}
Xin Wu, Danfeng Hong, and Jocelyn Chanussot.
\newblock Convolutional neural networks for multimodal remote sensing data classification.
\newblock \emph{IEEE Transactions on Geoscience and Remote Sensing}, 60:\penalty0 1--10, 2022.
\newblock \doi{10.1109/TGRS.2021.3124913}.

\bibitem[Xing and He(2023)]{xing_renewableenergy_2023}
Zhikai Xing and Yigang He.
\newblock Multi-modal multi-step wind power forecasting based on stacking deep learning model.
\newblock \emph{Renewable Energy}, 215:\penalty0 118991, 2023.
\newblock ISSN 0960-1481.
\newblock \doi{https://doi.org/10.1016/j.renene.2023.118991}.
\newblock URL \url{https://www.sciencedirect.com/science/article/pii/S0960148123008972}.

\bibitem[Yoon and Kang(2023)]{yoon_jpm_2023}
Taeyoung Yoon and Daesung Kang.
\newblock Multi-modal stacking ensemble for the diagnosis of cardiovascular diseases.
\newblock \emph{Journal of Personalized Medicine}, 13\penalty0 (2), 2023.
\newblock ISSN 2075-4426.
\newblock \doi{10.3390/jpm13020373}.
\newblock URL \url{https://www.mdpi.com/2075-4426/13/2/373}.

\bibitem[Ziadé et~al.(2023)]{flake8}
Tarek Ziadé et~al.
\newblock Flake8: Your tool for style guide enforcement, 2023.
\newblock URL \url{https://flake8.pycqa.org/}.

\bibitem[Zoabi et~al.(2021)Zoabi, Deri-Rozov, and Shomron]{zoabi_2021}
Yazeed Zoabi, Shira Deri-Rozov, and Noam Shomron.
\newblock Machine learning-based prediction of covid-19 diagnosis based on symptoms.
\newblock \emph{npj digital medicine}, 4\penalty0 (1):\penalty0 3, 2021.

\end{thebibliography}

\end{document}